\documentclass{article}

\PassOptionsToPackage{numbers}{natbib}
\usepackage[final]{neurips_2023}
\usepackage[utf8]{inputenc} % allow utf-8 input
\usepackage[T1]{fontenc}    % use 8-bit T1 fonts
\usepackage{hyperref}       % hyperlinks
\usepackage{url}            % simple URL typesetting
\usepackage{booktabs}       % professional-quality tables
\usepackage{amsfonts}       % blackboard math symbols
\usepackage{nicefrac}       % compact symbols for 1/2, etc.
\usepackage{microtype}      % microtypography
\usepackage{xcolor}         % colors

\usepackage{amsmath}
\usepackage{amssymb}
\usepackage{mathtools}
\usepackage{amsthm}

\usepackage{algorithm}
\usepackage{algorithmic}
\theoremstyle{plain}
\newtheorem{theorem}{Theorem}[section]

\newtheorem{lemma}[theorem]{Lemma}

\theoremstyle{definition}

\theoremstyle{remark}

\newcommand{\be}{\begin{equation}}
\newcommand{\ee}{\end{equation}}
\newcommand{\ben}{\begin{equation}}
\newcommand{\een}{\end{equation}}

\newcommand{\ds}{\displaystyle}

\newcommand{\Rn}{{\mathbb{R}^N}}

\newcommand{\R}{\mathbb{R}}

\newcommand{\kl}{\mathbb{KL}}

\title{Orthogonal Polynomials Approximation Algorithm:\\ a functional analytic approach to estimating probability densities}
\author{%
	Lilian W. Bia\l okozowicz\thanks{\href{mailto:lilianbialokozowicz+opaa@gmail.com}{Questions and comments are welcome, kindly send me an email.}} \\
	Vancouver, British Columbia, Canada
	}

\begin{document}

\maketitle

\begin{abstract}
	We present the new Orthogonal Polynomials Approximation Algorithm (OPAA), a parallelizable algorithm that estimates probability distributions using functional analytic approach: first, it finds a smooth functional estimate of a probability distribution, whether it is normalized or not; second, the algorithm provides an estimate of the normalizing weight; and third, the algorithm proposes a new computation scheme to compute such estimates.
	
	A core component of OPAA is a special transform of the square root of the joint distribution into a special functional space of our construct. Through this transform, the evidence is equated with the $L^2$ norm of the transformed function, squared. Hence, the evidence can be estimated by the sum of squares of the transform coefficients. Computations can be parallelized and completed in one pass.
	
	OPAA can be applied broadly to the estimation of probability density functions. In Bayesian problems, it can be applied to estimating the normalizing weight of the posterior, which is also known as the evidence, serving as an alternative to existing optimization-based methods.

\end{abstract}

\section{Introduction} 
Consider a distribution $p(\theta)$, which is a non-negative measurable function $p(\theta) \geq 0$ such that 
\ben
 \|p\| := \ds \int_\Rn p(\theta) d\theta < \infty.
\een We are interested in an approximation of $p(\theta)$, as well as its norm $\|p\|$ such that $p(\theta)/\|p\|$ is a probability density. In this paper, we introduce a new approximation approach, the \textbf{Orthogonal Polynomials Approximation Algorithm (OPAA)}. It estimates probabilities and their normalizing weights from a functional analytic perspective, can be parallelized, and the approximation can be arbitrarily close as the order of approximation increases.

The approximation of probability functions has many applications. In Bayesian problems, that arises in the estimation of the posterior: given a probability density function $P$ and a set of observations $X = (x_i)_{i=1}^{D}$, the posterior is
\ben
P(\theta|X) := \ds \frac{P(\theta, X)}{\int_\theta P(\theta, X) } 
\label{posterior}
\een and the evidence is
\ben
P(X) := \int_\theta P(\theta, X),
\label{evidence}
\een where $\theta = (\theta_i)_{i=1}^N$ is a set of (unknown) latent variables. There are many limitations that make it impractical to compute the posterior or evidence directly \footnote{We provide an example of the Mixed Gaussian Model in Appendix \ref{mixedGaussian} to illustrate this.}. For posterior inference, there are two major families of approach.

The first approach is random sampling, including Markov chain Monte Carlo methods such as the Metropolis--Hastings algorithm \citep{metropolis, hastings} and the Hamilton Monte Carlo algorithm \citep{hamiltonMC}. The second approach is the proxy model approach, which allows one to make inferences faster using the proxy instead. One example is variational inference, which was first developed about three decades ago \citep{peterson, hinton93, waterhouse, jordan99}. The idea behind variational inference is to find the optimal proxy of the posterior by means of optimization: first, one considers a family of proxies $q_\alpha$ with varying parameter $\alpha$. By means of descent methods, one tries to find the optimum parameters $\alpha*$ which minimize the ``closeness'' between $q_{\alpha*}(\theta)$ and the posterior. In the literature, one common choice to measure the "closeness" between the proxy and the actual posterior is the Kullback-Leibler Divergence ("KL-Divergence") \citep{kullbackLeibler}, defined as
\be
\kl(q||P(\theta|X)) := \ds \int q(\theta) \ds \log \left(\frac{q(\theta)}{P(\theta|X)} \right) d\theta .
\label{dklqp}
\ee

Note that the KL-divergence is asymmetric, meaning $\kl(q||p)$ is not necessarily the same as $\kl(p||q)$, thus giving rise to the choice between $\kl(q(\theta)||P(\theta|X))$ ("reverse KL") and $\kl(P(\theta|X)||q(\theta))$ ("forwards KL")\footnote{We refer the reader to Chapter 21 of \citet{mlBook} and Chapter 10 of \citet{bishop} for insightful discussions on the difference between the two.}. However, since it is hard to compute KL-divergence directly, the minimization problem is often transformed as follows
%substitute $P(\theta|X)=P(\theta, X)/P(X)$ into equation (\ref{dklqp}) and expand, which gives
\be
\log P(X) = \kl(q(\theta)|| P(\theta|X)) + \underbrace{\ds \int q(\theta) \ds \log \left( \frac{P(\theta,X)}{q(\theta)} \right) d\theta}_{L(q, X)} .
\label{elbo}
\ee

A number of important techniques to solve this minimization problem have evolved in the past three decades, such as the structured mean field approach \citep{saulJordan}, mean field approximation \citep{opper}, and stochastic variational inference \citep{svi}, with applications to streaming data \citep{theis} and anomaly detection \citep{soelch}. The reader may refer to \citet{bleiSurvey} for an in-depth survey of the development of variational inference techniques. However, among the algorithms developed, only a handful of them can solve for both the posterior and the evidence simultaneously. Also, most methods require some assumption of the prior distribution or the independence of the variables, but none of such assumptions is required for OPAA to work.

\subsection{Polynomials in Machine Learning} \label{opML}
 %and information theory \citep{dehesa, ahlswede}.

Polynomials have been a staple tool in the world of approximation, with applications spanning physics \citep{simonh, vinck}, random matrix theory \citep{deift} and statistics \citep{walter, diaconis}. However, applications of orthogonal polynomials to machine learning seem to be scarce, to the best of our knowledge. Perhaps the biggest obstacle to the application of orthogonal polynomials to higher-dimensional problems is the fact that there is no natural way to order them, hence the increasing computational complexity as the basis expands.

The reader may wonder the connections between OPAA and Polynomial Chaos Expansion (PCE), for both algorithms involve the use of polynomials. The truth is, they are completely different in nature, with the most critical difference being the fact that PCE uses a known prior to which the polynomials are orthogonal, while in OPAA polynomials are orthogonal with respect to measure of our choice (see \eqref{nu}). 

In particular, PCE appears in inverse modeling problems of which the goal is to obtain a model based on the observations, with the possibility that the model's parameters are outputs of random variables. Furthermore, in PCE, polynomials are chosen to be orthogonal with respect to the (known) joint distribution of the random variables and they are used to provide a simple estimation of the model (or the random variables). Advantages of this approach are that all quantities of inferential relevance come in closed forms, and the polynomial estimation of the posterior serves as a smooth proxy. However, knowledge of the prior is required for PCE to be applied, and that may not be available in certain problems.

Related to PCE is the work of \citet{nagelSudret}, who introduced \textit{Special Likelihood Expansion (SLE)}. The problem was set up as follows: let there be $N$ observations, $(y_n)_{n=1}^N$, unknown parameter $\theta \in D_\theta \subset \mathbb{R}^M$ that are generated by $M$ independent random variables $(X_i)_{i=1}^M$, and a forward model $F(\theta) \mapsto \hat{y} \in \mathbb{R}^N$. The authors proposed using orthogonal polynomials to estimate two things: first, the model $F$; second, the likelihood function $L(\theta) = P(Y| \theta)$, which the author called ``spectral likelihood'' and its representation by orthogonal polynomials, that is, ``spectral likelihood expansion (SLE)''. However, there are two major assumptions in \citet{nagelSudret} that limit its application to a broader class of problems: first, the polynomials are orthogonal with respect to a known prior $\pi(\theta)$; %which is the joint density of the parameters $\theta \in \mathbb{R}^M$;
second, it was assumed that the $M$ components of $\theta$ are independent. Such independence assumption allows the prior to be expressed as a product
\be
\pi(\theta) = \prod_{j=1}^M \pi_j (\theta_j),
\label{meanField}
\ee as in mean field techniques (see \citet{opper}). That, combined with the likelihood function $L(\theta)$ being expressed as a polynomial, implies that the posterior marginal of the $j$-th parameters has a simple closed expression (see equation (45) in the original paper)
\be
\pi(\theta_j| y) = \ds \frac{\pi_j(\theta_j)}{Z} \int L(\theta) \pi(\theta_{\sim j}) d\theta_{\sim j} ,
\ee owing to the orthogonality if the polynomials chosen with respect to the prior $\pi(\theta) d\theta$.

Essentially, the independence assumption of the latent variables means dropping all cross terms (also known as "interaction terms"). In the context of polynomial estimation, the immediate result is that the size of the basis grows linearly with the degree of polynomial estimation, instead of exponentially. For OPAA, that would mean the multivariate polynomials considered will be reduced to a subset of the form
\be
\Phi_\tau(\theta) = h_k(\theta_j).
\ee In other words, we only need to consider multi-indices of the form $\tau = (0, 0, \dots, k, 0, \dots, 0)$ which contain only one non-zero index. OPAA avoids such an assumption and it is up to the user to decide the degree of approximation.

Another notable application of orthogonal polynomials in machine learning was \citet{huggins}, known as the PASS-GLM algorithm (PASS stands for \textit{polynomial approximate sufficient statistics}. The PASS-GLM algorithm uses one-dimensional orthogonal polynomials of low degree to estimate the mapping function in Generalized Linear Models (GLMs), namely, given $N$ observations $(x_n, y_n)_{n=1}^N$ with $x_n \in \mathbb{R}^d$ and $y_n \in \mathbb{R}$, and parameter $\theta \in \mathbb{R}^d$, the following GLM was considered:
\be
\log p(y_{1:N}| x_{1:N}, \theta) = \sum_{n=1}^N \log p(y_n| g^{-1}(x_n \cdot \theta))% = \sum_{n=1}^N \phi(y_n, x_n\cdot \theta),
\ee where $g^{-1}(x_n \cdot \theta)$\footnote{This is often known as ``inverse mapping''. In the context of binary observations $y_n \in [0,1]$, the inverse mapping could be the logit function $g(x) = \log(x/1-x)$ for logistic regression, or $g(x) = \Phi^{-1}(x)$ with $\Phi$ being the standard Gaussian cumulative distribution function.} gives the expected value of $y_n$, and $\phi(y, s) := \log (y| g^{-1}(s))$ is the GLM mapping function which is to be estimated by orthogonal polynomials. In the special case of logistic regression, the mapping function is a function from $\mathbb{R}$ to $\mathbb{R}$ given by
\ben
\log p(y_n| x_n, \theta) = \phi_{\textrm{logit}}(y_n x_n \cdot \theta)
\label{logit}
\een where $\phi_{\textrm{logit}}(x) = - \log(1+e^{-s})$ is the logistic function. The PASS--GLM algorithm uses one-dimensional orthogonal polynomials up to degree $M$ ($M$ is small) to estimate the mapping function $\phi_{logit}(s)$ in equation (\ref{logit}). By replacing the mapping function with a low-order polynomial estimate (the author showed that $M=6$ sufficed even for $d \geq 100$), the PASS-GLM algorithm can quickly update the coefficients which give the approximate log-likelihood, which can then be used to construct an approximate posterior. One of their numerical experiments has further proven that the algorithm provides a practical tradeoff between computation and accuracy on a large data set (40 million samples with 20,000 dimensions). The algorithm is scalable because it only requires one pass over the data. The PASS--GLM algorithm has shown promise for the use of orthogonal polynomials in machine learning problems.
 
\section{Structure of the Paper}
This paper is structured as follows: in Section \ref{approxOP}, we present the new Orthogonal Polynomials Approximation Algorithm (OPAA), starting with the problem statement (Section \ref{statement}), an outline of the proof (Section \ref{outlineAlgo}), to be followed by the computation scheme (Section \ref{compScheme}) and a discussion of our contributions (Section \ref{ourContributions}).

In the remaining sections, we demonstrate why OPAA is valid mathematically. First, we state the two main theorems supporting OPAA (Section \ref{secProof}). Then we introduce the building blocks of the algorithm, including Hermite polynomials (Section \ref{hermite}), Gauss-Hermite quadrature in higher dimensions (Section \ref{ghq}), and how the OPAA computation scheme enables efficient computations by means of vector decomposition and parallelization (Section \ref{OPAAScheme}).
% Finally, OPAA's connections with other Bayesian problems and machine learning applications will be discussed (Section \ref{opML}).

%To supplement the discussion, we provide an example to illustrate how OPAA can be applied to a Mixed Gaussian Model (Appendix \ref{mixedGaussian}).

We provide a summary of the symbols and dimensions in Appendix \ref{appendixDi} for the convenience of the reader.

% auto-encoding variational Bayesian (AEVB) \citep{aevb}, variational Gaussian processes \citep{tranRB}. 

\section{The Orthogonal Polynomials Approximation Algorithm (OPAA)} \label{approxOP}
\subsection{Problem Statement}\label{statement}
Let there be $N$ variables, $\theta_1, \dots, \theta_N$ (or $\theta \in \Rn$) and a probability distribution $P(\theta)$. OPAA accomplishes three goals:
\begin{enumerate}
	\item It proves that there is a functional estimate of $P(\theta)$, meaning  
	\begin{equation}
		P(\theta) \approx \hat{P}_T(\theta) = \left( \sum_{\tau \in T} a_\tau \phi_\tau(\theta) \right)^2 \prod_{j=1}^N  e^{-\theta_j^2} ,
		% p_T(\theta)^2 
		\label{coeff1}
	\end{equation} where $\phi_\tau(\theta)$ are multivariate Hermite polynomials\footnote{The reader should note that OPAA can be generalized to other types of orthogonal polynomials, based on Theorem \ref{rieszTheorem}. However, we have found that Hermite polynomials seemed to be a good choice for practical purposes, because they have been implemented in popular computation libraries such as Scipy and Numpy.} and $a_\tau$ are real coefficients;
	\item The norm $\|P\| = \int P(\theta) d\theta$ can be estimated by the coefficients $a_\tau$ in \eqref{coeff1} above, that is,
	\begin{equation}
		\|P\| \approx \sum_{\tau} |a_\tau|^2 ;
	\end{equation}
	\item The coefficients $a_\tau$ are given by the following formula and can be estimated by quadrature,
	\begin{equation}
		a_\tau := \int_{\Rn} P(\theta)^{1/2} \phi_\tau(\theta) \left(\prod_{j=1}^N e^{-\theta_j^2/2} \right) d\theta_{1:N} .
	\end{equation}
\end{enumerate} 

A useful result that follows from OPAA is the following:
\begin{lemma}
	Every probability density $f(\theta)$ on $\mathbb{R}$ (that is, $f(\theta) \geq 0$ and $\int f(\theta) d\theta=1$) can be estimated as follows
	\begin{equation}
		f(\theta) \approx \hat{f}_J(\theta):= \ds \frac{\left(\sum_{j=0}^J a_j h_j(\theta)\right)^2}{\sum_{j=0}^J |a_j|^2} e^{-\theta^2} ,
	\end{equation} where $h_j(\theta)$ are Hermite functions of degree $j$.
\end{lemma} An in-depth discussion of Hermite polynomials is provided for clarity in Section \ref{hermite}).

\subsection{Outline of the Algorithm} \label{outlineAlgo}
We consider the functional space $L^2(d\nu_{1:N})$ associated with the following measure on $\Rn$
\ben
d\nu_{1:N}(\theta) := \prod_{j=1}^N e^{-\theta_j^2} d\theta_j ,
\label{nu}
\een where $\theta = (\theta_1, \theta_2, \cdots, \theta_N) \in \Rn$. Let $h_i(x)_{i=0}^\infty$ be the normalized one-dimensional Hermite polynomials that are orthogonal with respect to the measure $d\nu = e^{-\theta^2} d\theta$ on $\R$, that is,
\be
\int_\R h_i(x) h_j(x) e^{-\theta^2} dx = \begin{cases} 1 & \text{if  } i=j; \\ 0 & \text{otherwise.} \end{cases}
\label{orthogonality}
\ee Such orthogonality implies that the tensor products of Hermite polynomials of the form
\be
\phi_\tau(\theta) := h_{i_1}(\theta_1) h_{i_2}(\theta_2) \cdots h_{i_N}(\theta_N)
\label{representation}
\ee form an orthogonal polynomial basis that is orthogonal with respect to $d\nu_{1:N}$. The $N$-tuple $\tau = (i_1, i_2, \dots, i_N)$ is known as a \textbf{multi-index}, with each index $i_k$ being a non-negative integer. The \textbf{degree} of this multi-index is given by $|\tau| := \sum_{j=1}^{N} i_j$.

The measure $d\nu_{1:N}$ is special because it fulfills the finite moment criterion (\ref{finiteMoments}). Hence, by the Riesz Theorem (Theorem \ref{rieszTheorem}), the family of polynomials is dense in $L^2(d\nu_{1:N})$ (some call it a ``complete basis''). In particular, we let
\ben
\tilde{P}(\theta) := P(\theta) \prod_{j=1}^N e^{\theta_j^2}.
\label{PtildeDef}
\een Observe that $\tilde{P}^{1/2}$ is in $L^2(d\nu_{1:N})$ because 
\be
  \left(\int_\Rn \left |\tilde{P}(\theta)^{1/2} \right|^2 d\nu_{1:N} \right)^{1/2}  = \left(\int_\Rn |P(\theta)| d\theta_{1:N} \right)^{1/2} = \|P\|^{1/2} < \infty.
\label{px}
\ee

Next, we transform $\tilde{P}(\theta)^{1/2}$ into an infinite series by projecting it onto the polynomial basis $(\phi_\tau)_{\tau}$. The transform coefficients are given by\footnote{To understand intuitively why the integral \eqref{atauDef} is well defined, the reader may refer to Appendix \ref{secHF}.} 
\ben
a_\tau := \int_{\Rn} P(\theta)^{1/2} \phi_\tau(\theta) \left(\prod_{j=1}^N e^{-\theta_j^2/2} \right) d\theta_{1:N} .
\label{atauDef}
\een

The density of polynomials in $L^2(d\nu_{1:N})$ allows one to invoke the Parseval Identity, which gives
\ben
\|\tilde{P}(\theta)^{1/2}\|_{L^2(d\nu_{1:N})}^2 = \sum_{\tau} |a_\tau|^2.
\label{parseval1}
\een Combining this with (\ref{px}), we obtain
\ben
\|P\| = \sum_{\tau} |a_\tau|^2 .% \approx \sum_\tau |\hat{a}_\tau|^2 .
\label{parsevalEst}
\een The fact that the coefficients $(a_\tau)_\tau$ are absolutely convergent implies that the summation can be executed in any order.

Furthermore, for a finite set of multi-indices $T$ such that $\sum_{\tau \in T} |a_\tau|^2 \approx \|P\|$, if we define the polynomial
\ben
p_T(\theta) := \left( \sum_{\tau \in T} |a_\tau|^2 \right)^{-1/2} \left( \sum_{\tau \in T} a_\tau \phi_\tau(\theta) \right);
\label{pT}
\een then 
\be
\pi_{1:N}(\theta) := p_T(\theta)^2 \prod_{j=1}^N e^{-\theta_j^2}
\label{pTDef}
\ee is a density (meaning it is non-negative and $\int \pi_{1:N}(\theta) d\theta = 1$) and that 
\ben
\pi_{1:N}(\theta) \approx P(\theta)/\|P\|.
\een In fact, any non-zero polynomial of the form (\ref{pT}) will have norm $1$ and the corresponding $\pi_{1:N}$ will be a probability density function.

In fact, $p_T(\theta)$ in \eqref{pT} can be expressed as a product of Hermite functions, which are defined as
\be
\psi_n(x) := h_n(x) e^{-x^2/2}, n \geq 0.
\ee We provide a brief discussion of Hermite functions in Appendix \ref{secHF}.
% where 
%\be
%\int_\Rn \pi_{1:N}(\theta) d\theta = <p_T, p_T>_{d\nu_{1:N}}^2 = 1.
%\ee

\subsection{Outline of the Computation Scheme} \label{compScheme}
It has been observed that random sampling methods applied to integrals such as (\ref{atauDef}) may result in high variance. For that reason, we propose the use of Gauss--Hermite quadrature to estimate $a_\tau$ (more details in Section \ref{OPAAScheme}). We will demonstrate that not only does this provide a more reliable estimation, it also enables us to compute the values fast.

First, we choose a quadrature order $\Gamma$. Quadrature of order $\Gamma$ works well to approximate function which can be well estimated by a polynomial of degree $2\Gamma-1$. For that reason, usually a single-digit $\Gamma$ will suffice.

From the one-dimensional quadrature nodes and weights $({r}_i, {w}_i)_{i=1}^\Gamma$\footnote{These constants are available in Numpy libraries and numerical analysis handbooks.}, we form our multivariate nodes in $\Rn$, $(\tilde{r}^{(j)})_{(j)}$; and weights, $(\tilde{w}^{(j)})_{(j)}$ for each grid-index $(j) \in I$.

The transform coefficients can then be estimated by
\ben
a_\tau \approx \sum_{(j) \in I} \tilde{w}_{(j)} P(\tilde{r}^{(j)})^{1/2} \phi_\tau(\tilde{r}^{(j)}).
\label{decompose}
\een

The right hand side of \eqref{decompose} can be expressed as\footnote{The symbols $\odot$ and $\cdot$ denote the pointwise multiplication and dot product of two vectors respectively.}
\ben
a_\tau \approx %\vec{\Pi} \cdot \vec{\Phi}_\tau 
\vec{W} \odot \vec{P} \cdot \vec{\Phi}_{\tau},
\label{discretization}
%	\vec{\Pi} & := &  [\tilde{w}^{(j)}]_{(j)} \odot [P({\tilde{r}}^{(j)}, X)^{1/2}]_{(j)} \label{PiDef} \\
%\vec{\Phi}_{\tau} & := & C .  \label{PhiDef}
\een This decomposition into three vectors will bring many computational advantages that help tackle the problem of dimensionality, with the major advantages being: (1) most of the values can be obtained from simple arithmetic, and (2) both $\vec{W}$ and $\vec{\Phi}_\tau$ depend on values of size $O(\Gamma \cdot d)$ where $d$ is the degree of polynomial estimation. The details will be provided in Section \ref{OPAAScheme}, after we introduce the building blocks of the algorithm. Besides, the expression \eqref{discretization} allows parallelization, which substantially increases the speed of computation.

\begin{algorithm}[tb]
	\caption{The Orthogonal Polynomials Approximation Algorithm (OPAA)}
	\label{algo1}
	\begin{algorithmic}
		\STATE \textbf{Input} Quadrature order $\Gamma$. Joint distribution $P(\theta)$.
		\STATE \textbf{Output} Coefficients $(a_\tau)_{\tau \in T}$ which can be used to compute the weight $\|P\|$ and a smooth probability density function $f_T(\theta)$ that estimates $P(\theta)$.
		\WHILE{$\Sigma = \sum_{|\tau| < d} |a_\tau|^2$ does not converge}
		\STATE Increase the degree $d$ by $1$.
		\STATE Compute $h_{d}(x)$ for $x = \tilde{r}_1, \dots, \tilde{r}_\Gamma$.
		\FOR{multi-index $\tau$ of degree $d$}
		\STATE Compute $\Phi_\tau = (\phi_\tau(\tilde{r}^{(j)}))_{(j) \in I}$ \COMMENT{Eq. (\ref{PhiDef})}
		\STATE Compute $\hat{a}_\tau = \vec{W} \odot \vec{P} \cdot \vec{\Phi}_\tau$ \COMMENT{Eq. (\ref{atauVector})}
		%Compute ${a}_\tau = \vec{\Pi} \cdot \Phi_\tau$ 
		\STATE Add ${a}_\tau^2$ to $\Sigma$.
		\ENDFOR
		
		\ENDWHILE \\
		%\RETURN $(a_\tau)_{\tau}$
	\end{algorithmic}
\end{algorithm}

\subsection{Our Contributions} \label{ourContributions}
%Our contributions can be categorized as follows:

\textbf{A new functional analytic perspective.} Instead of finding a proxy through optimization, we identified a functional transform which allows the decomposition of the density function into a series of complete basis functions. It is important to note that the choice of basis function is critical, for the lack of completeness of this basis will cause a key equality \eqref{parsevalEst} to fail, and the whole argument will fall apart.

\textbf{No assumption about knowledge of the prior or independence of the variables.} OPAA does not assume any knowledge of the prior or the independence of the latent variables, a common assumption to simplify the computation as all cross terms will be annihilated.

%\item 
\textbf{An accurate, parallelizable and efficient computation scheme.} The OPAA computation scheme brings a few advantages. By using quadrature, it counters the variance problems from random sampling methods; the discretization of $a_\tau$ in \eqref{discretization} allows for efficient computation. In particular, both $\vec{W}$ and $\vec{\Phi}_\tau$ are independent of the distribution in question $P$, so both $\vec{W}$ and $\vec{\Phi}_\tau$ are essentially universal constants that apply to all OPAA applications. Besides, $\vec{W}$ only depends on $\Gamma$ quadrature weights; and $\vec{\Phi}_\tau$ on a set of $\Gamma \cdot |\tau|$ values, namely,
\be
V_{|\tau|} := \{h_d(\tilde{r}_i): 0 \leq d \leq |\tau|; 1 \leq i \leq \Gamma \}.
\label{VdDef}
\ee 

OPAA can produce arbitrarily good approximations as we increase the degree of polynomial approximation and order of quadrature $\Gamma$.

\section{Supporting Theorems in Functional Analysis} \label{secProof}
To illustrate the mathematical soundness of OPAA, we output the two classical results in functional analysis which were invoked:
\begin{enumerate}
	\item The Riesz Theorem, which guarantees the density of polynomials in $L^2(d\nu_{1:N})$ and hence the Parseval equality $\|P\| = \sum_{\tau} |a_\tau|^2$ (see \eqref{parsevalEst});
	 %if the measure $d\nu_{1:N}$ satisfies the moment condition \eqref{finiteMoments};
	\item The decomposition of $P(\theta)$ using a complete set of basis functions (see Theorem \ref{mainTheorem}).
\end{enumerate}

First, we state the Riesz's Theorem, a classic result in approximation theory.
\begin{theorem}[Density of polynomials in $L^2$]\citep{riesz}.  \label{rieszTheorem}
Let $\nu$ be a measure on $\mathbb{R}^N$ satisfying
\ben
\ds \int_\Rn e^{c|\theta|} d\nu < \infty
\label{finiteMoments}
\een for some constant $c>0$, where $|\theta| = \sum_{j=1}^{N} |\theta_j|$; then the family of polynomials is dense in $L^2(\nu)$. In other words, given any $f \in L^2(\nu)$, there is a sequence of polynomials $f_n(\theta)$ such that 
\be
\lim_{n \to \infty} \int_\Rn |f(\theta) - f_n(\theta)|^2 d\nu = 0 .
\ee
\end{theorem}

Criterion (\ref{finiteMoments}) implies that all polynomials are in $L^p(\nu)$, for any $p \geq 1$. To see that, it suffices to show that for any $c>0$ and integer $k \geq 0$
\be
\lim_{x \to \infty} \ds \frac{x^k}{e^{cx}} < \infty.
\ee This could be proven by the repeated application of the L'H\^{o}pital rule. Related moment problems are discussed in depth by \citet{akhiezer} (Theorem 2.3.3 and Corollary 2.3.3). A nice short proof of the result was presented in \cite{schmuland}.

Next, we state a classic result in functional analysis.
%To apply it to prove OPAA, just set $F(\theta) = \tilde{P}(\theta)^{1/2}$ and $\mu = \nu_{1:N}$ (defined in \eqref{PtildeDef} and \eqref{nu} respectively).
 
\begin{theorem} (Decomposition of a function into a series formed by a complete basis) \label{mainTheorem} Given a measure $\mu$ on $\Rn$ and a function $F(\theta)$ which is in $L^2(\nu)$, meaning
\be
\int_{\Rn} |F(\theta)|^2 d\mu(\theta) < \infty.
\ee	If $(\phi_\tau)_\tau$ is a complete basis in $L^2(\mu)$,  the function $F(\theta)$ can be estimated by functions of the form
\be
f_n(\theta) := \sum_{j=0}^{n} \sum_{|\tau| = j} a_\tau \phi_\tau(\theta),
\ee and the coefficients $a_\tau$ are given by 
\ben
a_\tau = \int_\Rn F(\theta) \phi_\tau(\theta) d\mu(\theta)
\label{coefficients}
\een Moreover, the following equality holds
\ben
\|F\|_{L^2(\mu)} = \int_{\Rn} |F(\theta)|^2 d\mu(\theta) = \left(\sum_{\tau} |a_\tau|^2 \right)^{1/2}.
\label{L2norm}
\een
\end{theorem}

Most readers may be familiar with the Fourier Transform on $[0, 2\pi]$ or Taylor expansions. Theorem \ref{L2norm} can be vaguely understood as a generalization of those. However, we would like to remind the reader that when results are extended to an unbounded domain like $\Rn$, generalizations are not straightforward. Equation \ref{L2norm}, for instance, may become an inequality if the basis is not complete (known as the Bessel's inequality) and a major piece of OPAA (that is, equation \eqref{parsevalEst}) would fail.

\section{Building Blocks of the Algorithm} \label{buildingBlocks}
%In this section, we provide an in-depth discussion of the major building blocks of OPAA, which are Hermite Polynomials (Section \ref{hermite}), Gauss--Hermite Quadrature in one-dimension (Section \ref{ghq}) and its extension to higher dimensions (Section \ref{ghqhd}). After introducing these preliminaries, we will illustrate how OPAA reduces the complexity of the computation.

\subsection{Hermite Polynomials and Density of Polynomials}\label{hermite}
Hermite polynomials\footnote{The Hermite polynomials used in this paper are often known as the physicists' Hermite polynomials because they are orthogonal to $e^{-x^2}$ instead of $e^{-x^2/2}$. The reader should be careful when coding up the results, because both sets of polynomials are usually available in standard numerical analysis libraries.} are polynomials\ on $\R$ that are orthogonal with respect to the measure 
\be
d\nu := e^{-x^2}dx \text{ on } \R.
\ee

%The inner product $<, >_\nu$ and $\|\cdot \|_\nu$ for the space $L^2(\nu(\R))$ are defined respectively as
%\be <f, g>_\nu := \int_\R f(x) g(x) d\nu \text{ and } \|f\|_\nu := <f, f>_\nu^{1/2} . \ee In particular, the

Hermite polynomials satisfy the following orthogonality relation
\ben \ds \int_\R H_m (x) H_n(x) d\nu(x) = \sqrt{\pi} 2^n n! \delta_{nm} .
\label{innerProduct}
\een Normalized Hermite polynomials are denoted as $h_n(x):=H_n(x)/\|H_n\|$. The Hermite polynomials used in this paper are
\begin{eqnarray}
	H_0(x) = 1, &  h_0(x) = \pi^{-1/4} \\
	H_1(x) = 2x, & h_1(x) = \sqrt{2} \pi^{-1/4} x
\end{eqnarray} and the higher order polynomials can be obtained from the following recurrence relation
\be
	H_{n+1}(x) = 2x H_n(x) - 2n H_{n-1}(x).
	\label{recurrence}
\ee
The measure $\nu$ is the building block of the measure $\nu_{1:N}$ defined in equation \eqref{nu}. A critical property of $\nu$ is that it has finite moments, that is, there is a constant $c>0$ such that
\be
\int_\R e^{c|\theta|} d\nu %= 2 \int_{\theta \geq 0} e^{-\theta^2+ c\theta} d\theta
\leq 2 \int e^{{-(\theta-c/2)^2 + \frac{c^2}{4}}} d\theta < \infty .
\ee Following a similar argument, one can prove that
\ben
\int_\Rn e^{c |\theta|} d \nu_{1:N}(\theta) = \ds \prod_{j=1}^{N} \int e^{c |\theta_j|} e^{-\theta_j^2} d\theta_j < \infty .
\label{measureRequirement}
\een Condition (\ref{measureRequirement}) makes $\nu_{1:N}$ eligible for the Riesz Theorem (Theorem \ref{rieszTheorem}), which implies the density of polynomials in $L^2(\nu_{1:N})$. Without this density, the equality (\ref{parsevalEst}) may not hold.

Apart from Hermite polynomials, Chebyshev's polynomials and Jacobi polynomials are among the most commonly known families of orthogonal polynomials. For a comprehensive introduction to orthogonal polynomials, the readers may refer to \citet{simon, koornwinder}.

\subsection{Gauss--Hermite Quadrature in One Dimension} \label{ghq}
We start with a preview of Gauss--Hermite quadrature in one dimension. If the function $f(x)$ can be estimated by polynomials of degree $2\Gamma-1$ or less ($\Gamma$ for ``Gaussian''),
\ben
\int_\R f(x) e^{-x^2}dx \approx \sum_{i=1}^\Gamma w_i f(r_i)
\label{intf}
\een where the nodes $r_1, {r}_2, \dots, {r}_\Gamma$ are the distinct roots of $H_\Gamma(x)$; and the weights ${w}_1, {w}_2, \dots, {w}_\Gamma$ are given by
\ben
{w}_i:= \ds \frac{1}{\Gamma h_{\Gamma-1}({r}_i)^2}, 1 \leq i \leq \Gamma .%\ds \frac{2^{n-1} n! \sqrt{\pi}}{n^2 H_{n-1}(r_i)^2} = 
\label{weightsDef}
\een

Given that in equation (\ref{atauDef}), we are integrating against $e^{-\theta^2/2}$, we will need to perform a simple change of variables: we define 
\begin{eqnarray}
\tilde{w}_i & := & \sqrt{2}{w}_i, \label{weightDef}\\
\tilde{r}_i & := & \sqrt{2} {r}_i. \label{nodeDef}
\end{eqnarray} The quadrature formula now becomes
\begin{equation}
\int_\R f(\theta) e^{-\theta^2/2} d\theta  = \sqrt{2} \int f(\sqrt{2}\theta) e^{-\theta^2} d\theta \approx \sum_{i=1}^\Gamma {\tilde{w}_i} f(\tilde{r}_i) .
\label{quadApprox}
\end{equation} Furthermore, (\ref{quadApprox}) turns into an equality if $f(\theta)$ is a polynomial of degree $\leq 2\Gamma-1$. To ensure that the constants used in the computation are correct, we encourage the reader to check if setting $f=1$ returns $\sum_{i}^\Gamma \tilde{w}_i = \int_\R e^{-\theta^2/2} = \sqrt{2 \pi}$.

Note that $w_i$ in (\ref{weightsDef}) is well defined because $h_{\Gamma-1}(r_i)\not = 0$, which follows from the interlacing zeros properties of orthogonal polynomials, implying that $h_{\Gamma-1}$ and $h_\Gamma$ do not share any zeroes.

The quadrature nodes and weights, $r_i$ and $w_i$, are constants that are available in handbooks for numerical analysis (for example, \cite{abramowitz, olver}) and Numpy library\footnote{\url{https://numpy.org/doc/stable/reference/generated/numpy.polynomial.hermite.hermgauss.html}} for they naturally appear in many problems in numerical analysis. The reader may refer to  \citet{cools} for a survey of the field.

\subsection{Quadrature In Higher Dimensions} \label{ghqhd}
We tried to yield to the literature to establish a stable estimation of $a_\tau$, an integral in $\Rn$. However, we are not aware of any general scheme of computation in the literature for $\Rn$ \citep{lu, millan, zandt}. To the best of our knowledge, the numerical results are mostly restricted to low dimensions or to specific functions. For the aforementioned reasons, we decided to expand into $\Rn$ based on equation (\ref{intf}), expecting ample room for improvements in the future.

We consider an integer $\Gamma$, the order of quadrature of our choice. First, observe that for any fixed $\theta_2, \dots, \theta_N$, $f$ can be treated as a function of $\theta_1$ such that
\be
\int_\R f(\theta_1, \theta_2, \dots, \theta_N) e^{-\theta_1^2/2}d\theta_1  = \ds \sum_{j=1}^\Gamma \tilde{w}_j f(\tilde{r}_j,\theta_2, \dots, \theta_n) .
\ee

Furthermore, by the independence of the quadrature weights $w_i$ and the nodes $r_i$ with respect to $f$, we can proceed inductively and obtain
\begin{multline}
	\label{highDim}
\ds \int_\Rn f(\theta_1, \dots, \theta_N) \prod_{j=1}^N e^{-\theta_i^2/2} d\theta \\ \approx \int \sum_{i=1}^\Gamma \tilde{w}_i P(\tilde{r}_i, \theta_2, \dots, \theta_n)  \prod_{j=2}^\Gamma e^{- \theta_j^2/2} d\theta_2 \cdots d\theta_N \\
 = \sum_{i_1=1}^\Gamma \dots   \sum_{i_N=1}^\Gamma \tilde{w}_{i_1} \dots \tilde{w}_{i_N} f(\tilde{r}_{i_1}, \dots, \tilde{r}_{i_N}) .
\end{multline}

In other words, in higher dimensions, the index set becomes
\be
I = [1, 2, \dots, \Gamma]^N
\label{indexsetDef} 
\ee Each element in $I$ is known as a \textbf{grid index} and comes in the form
\be
(j) = (j_1, j_2, \dots, j_N),
\ee with each index $1 \leq j_k \leq \Gamma$. For such $(j)$, the corresponding \textbf{nodes} and \textbf{weights} for high-dimensional quadrature are
\begin{eqnarray}
	{\tilde{r}}^{(j)} & := &  (\tilde{r}_{j_1}, \tilde{r}_{j_2}, \dots, \tilde{r}_{j_N}). \label{rtildeDef}  \\ %useful
	{\tilde{w}}^{(j)} & := & \prod_{k=1}^N \tilde{w}_{j_k}, \label{wtildeDef}
\end{eqnarray} where $\tilde{w}_k$ and $\tilde{r}_k$ are defined in equations \eqref{weightDef} and \eqref{nodeDef} respectively.

\subsection{Vectorization and Parallelization}  \label{OPAAScheme}
Applying results from the previous section to $a_\tau$ in equation \eqref{atauDef}, we obtain
\ben
%a_\tau \approx \sum_{r^{(j)}, w^{(j)}} w^{(j)} P(r^{(j)}, X)^{1/2} \phi_\tau(r^{(j)}).
a_\tau \approx \hat{a}_\tau := \sum_{(j) \in I} {\tilde{w}}^{(j)} P({\tilde{r}}^{(j)})^{1/2} \phi_\tau({\tilde{r}}^{(j)}) .
\label{atau1}
\een

The right hand side of equation (\ref{atau1}) can be expressed in vector form
\ben
\hat{a}_\tau = \vec{W} \odot \vec{P} \cdot \vec{\Phi}_\tau ,
\label{atauVector}
\een where 
\begin{eqnarray}
	\vec{W} & = & [\tilde{w}^{(j)}]_{(j)}, \label{Wdef} \\
	%\vec{\Pi} & := &  [\tilde{w}^{(j)}]_{(j)} \odot [P({\tilde{r}}^{(j)}, X)^{1/2}]_{(j)} \label{PiDef} \\
	\vec{P} & = & [P({\tilde{r}}^{(j)})^{1/2}]_{(j)}, \label{PDef} \\
   \vec{\Phi}_{\tau} & := & [\phi_\tau({\tilde{r}}^{(j)})]_{(j)} .  \label{PhiDef}
\end{eqnarray}

This decomposition into three vectors plays a crucial role in reducing the computation complexity because of the following facts:
\begin{enumerate}
	\item Out of the three vectors, both $\vec{\Phi}_\tau$ and $\vec{W}$ are independent of the probability function $P(\theta)$ in question. In other words, they are universal constants for OPAA that can be applied as the number of observations increases, or for a totally different problem.
	
	\item When starting from scratch, one can compute $\vec{\Phi}_\tau$ inductively in increasing degrees of $\tau$. As the degree of estimation increases by $1$ (say, from $k-1$ to $k$), one only needs to compute $\Gamma$ new values, namely,
	\be
	\{ h_k(\tilde{r}_j) \text{ for } 1 \leq j \leq  \Gamma \}.
	\label{depVals}
	\ee In fact, $\vec{\Phi}_\tau$ only depends on $\Gamma \cdot |\tau|$ values, which are
	\be
	\{h_k(\tilde{r}_j) \text{ for } 1 \leq j \leq  \Gamma, 0 \leq k \leq |\tau| \}.
	\ee 
	
	\item Due to the fact that the weights $\tilde{w}^{(j)}$ are formed by multiplications, there are many repetitions: for example, if $\Gamma=5$ and $N=10$, the following two grid-indices
	\begin{eqnarray*}
	(j_1) & = & [1, 2, 1, 4, 5, 1, 3, 1, 2, 3] \\
	(j_2) & = & [5, 1, 2, 1, 3, 1, 1, 2, 3, 4]
	\end{eqnarray*} result in the same weight. That is because by definition,
\be
\tilde{w}^{(j_1)} = \tilde{w}^{(j_2)}
%= w_1 \cdot w_2 \cdot w_1 \cdot w_4 \cdot w_5 \cdot w_1 \cdot w_3 \cdot w_1 \cdot w_2 \cdot w_3 
= w_1^{4} \cdot w_2^2 \cdot w_3^2 \cdot w_4 \cdot w_5.
\ee In fact, the size of the index set $I$ is $9765625$ while there are only $1001$ possible distinct values in the set of weights. It only took $11.7$ seconds on a personal computer to compute $\tilde{w}^{(j)}$ for all grid-indices $(j)$ in this index set.
\end{enumerate}
%Figure \ref{p1} shows the computation times for various $\Gamma$'s. It demonstrates how parallelization significantly reduced the computation times.

%\begin{figure}[ht]
%	\vskip 0.2in
%	\begin{center}
%		\centerline{\includegraphics[width=\columnwidth]{computation_time_weights}}
%		\caption{Computation times for $(\tilde{w}^{(j)})_{(j)}$ with $\Gamma=3, 4, 5$, with and without parallelization. We performed the computations on a 6-core CPU.}
%		\label{p1}
%	\end{center}
%	\vskip -0.2in
%\end{figure}
%\end{enumerate}
%\subsection{Special Case: Independent Latent Variables.} \label{independent}
%In the case that all latent variables are known to be independent, the algorithm reduces to a simpler form, as we no longer need to consider any cross terms. As a result, the basis functions are those that can be represented by a multi-index $\tau$ in the form $(0,\dots, 0, k, 0, \dots, 0)$ (that is, only one non-zero integer $k$ in the $m$-th direction) and the corresponding polynomial is
%\be \phi_\tau(\theta) = h_k(\theta_m) . \ee

\section{Conclusions}
OPAA is an efficient algorithm to give functional estimates of probability functions and their normalizing weights. In Bayesian problems, by treating the evidence as the functional norm (see (\ref{parsevalEst})), the problem is transformed into the computation of estimation coefficients $a_\tau$. 

By leveraging Gauss--Hermite quadrature, the coefficients $a_\tau$ can be decomposed into a product of three vectors (Section \ref{OPAAScheme}), two of them ($\vec{W}$ and $\vec{\Phi}_\tau$) are universal constants for OPAA that are independent of the problem; and they depend on values that are of size $O(\Gamma \cdot |\tau|)$, where $\Gamma$ is the order of quadrature and $|\tau|$ is the degree of polynomial estimation.

\section*{Acknowledgements}
The author would like to thank Professor Evans Harrell for helpful discussions.

%Imports the bibliography file "root.bib"
\vskip 0.2in

\bibliography{biblio}

\begin{thebibliography}{35}
\providecommand{\natexlab}[1]{#1}
\providecommand{\url}[1]{\texttt{#1}}
\expandafter\ifx\csname urlstyle\endcsname\relax
  \providecommand{\doi}[1]{doi: #1}\else
  \providecommand{\doi}{doi: \begingroup \urlstyle{rm}\Url}\fi

\bibitem[Abramowitz \& Stegun(1972)Abramowitz and Stegun]{abramowitz}
Abramowitz, M. and Stegun, I.~A.
\newblock \emph{Handbook of Mathematical Functions}.
\newblock Dover, 1972.

\bibitem[Akhiezer(1965)]{akhiezer}
Akhiezer, N.~I.
\newblock \emph{The Classical Moment Problem and Some Related Questions in
  Analysis}.
\newblock Dover Publications, 1965.

\bibitem[Bishop(2006)]{bishop}
Bishop, C.
\newblock \emph{Pattern Recognition and Machine Learning}.
\newblock Springer New York, 2006.

\bibitem[Blei et~al.(2017)Blei, Kucukelbir, and MacAuliffe]{bleiSurvey}
Blei, D.~M., Kucukelbir, A., and MacAuliffe, J.~D.
\newblock Variational inference: A review for statisticians.
\newblock \emph{Journal of the American Statistical Association}, 112:\penalty0
  859--877, 2017.
\newblock \doi{https://doi.org/10.1080/01621459.2017.1285773}.

\bibitem[Cools(2002)]{cools}
Cools, R.
\newblock Advances in multidimensional integration.
\newblock \emph{Journal of Computational and Applied Mathematics}, 149\penalty0
  (1):\penalty0 1--12, 2002.

\bibitem[Deift(2000)]{deift}
Deift, P.
\newblock \emph{Orthogonal Polynomials and Random Matrices: A Riemann-Hilbert
  Approach}, volume~3.
\newblock Courant Lecture Notes. American Mathematical Society, 2000.

\bibitem[Diaconis et~al.(2008)Diaconis, Khare, and Saloff-Coste]{diaconis}
Diaconis, P., Khare, K., and Saloff-Coste, L.
\newblock Gibbs sampling, exponential families and orthogonal polynomials.
\newblock \emph{Statistical Science}, 23\penalty0 (2):\penalty0 151--178, 2008.
\newblock \doi{https://doi.org/10.1214/07-STS252}.

\bibitem[Hastings(1970)]{hastings}
Hastings, W.~K.
\newblock Monte carlo sampling methods using markov chains and their
  applications.
\newblock \emph{Biometrika}, 57\penalty0 (1):\penalty0 97--103, 1970.

\bibitem[Hille(1926)]{hille}
Hille, E.
\newblock A class of reciprocal functions.
\newblock \emph{The Annals of Math.}, 27:\penalty0 427--464, 1926.

\bibitem[Hinton \& Camp(1993)Hinton and Camp]{hinton93}
Hinton, G. and Camp, D.~V.
\newblock Keeping the neural networks simple by minimizing the description
  length of the weights.
\newblock \emph{Computational Learning Theory}, pp.\  5--13, 1993.

\bibitem[Hoffman \& Gelman(2014)Hoffman and Gelman]{hamiltonMC}
Hoffman, M.~D. and Gelman, A.
\newblock The no-u-turn sampler: Adaptively setting path lengths in hamiltonian
  monte carlo.
\newblock \emph{Journal of Machine Learning Research}, 15:\penalty0 1593--1623,
  2014.

\bibitem[Hoffman et~al.(2013)Hoffman, Blei, Wang, and Paisley]{svi}
Hoffman, M.~D., Blei, D.~M., Wang, C., and Paisley, J.
\newblock Stochastic variational inference.
\newblock \emph{Journal of Machine Learning Research}, 14:\penalty0 1307--1347,
  2013.

\bibitem[Huggins et~al.(2017)Huggins, Adams, and Broderick]{huggins}
Huggins, J.~H., Adams, R.~P., and Broderick, T.
\newblock Pass-glm: polynomial approximate sufficient statistics for scalable
  bayesian glm inference.
\newblock \emph{Proceedings of the 31st Annual Conference on Neural Information
  Processing Systems (NIPS 2017)}, 2017.

\bibitem[Jordan et~al.(1999)Jordan, Ghahramani, Jaakkola, and Saul]{jordan99}
Jordan, M.~I., Ghahramani, Z., Jaakkola, T., and Saul, L.
\newblock Introduction to variational methods for graphical models.
\newblock \emph{Machine Learning}, 37:\penalty0 183--233, 1999.

\bibitem[Koornwinder(2013)]{koornwinder}
Koornwinder, T.
\newblock Orthogonal polynomials, a short introduction.
\newblock In \emph{C. Schneider, J. Bluemlein J. (eds) Computer Algebra in
  Quantum Field Theory. Texts \& Monographs in Symbolic Computation (A Series
  of the Research Institute for Symbolic Computation, Johannes Kepler
  University, Linz, Austria)}, pp.\  145--170. Springer, Vienna, 2013.

\bibitem[Kullback \& Leibler(1951)Kullback and Leibler]{kullbackLeibler}
Kullback, S. and Leibler, R.
\newblock On information and sufficiency. the annals of mathematical
  statistics.
\newblock \emph{The Annals of Mathematical Statistics}, 22\penalty0
  (1):\penalty0 79--86, 1951.

\bibitem[Lu \& Darmofal(2004)Lu and Darmofal]{lu}
Lu, J. and Darmofal, D.~L.
\newblock Higher-dimensional integration with gaussian weight for applications
  in probabilistic design.
\newblock \emph{SIAM Journal of Scientific Computing}, 26\penalty0
  (2):\penalty0 613--624, 2004.

\bibitem[Metropolis et~al.(1953)Metropolis, Rosenbluth, Rosenbluth, and
  Teller]{metropolis}
Metropolis, N., Rosenbluth, A.~W., Rosenbluth, M.~N., and Teller, A.~H.
\newblock Equation of state calculations by fast computing machines.
\newblock \emph{The Journal of Chemical Physics}, 21, 1953.

\bibitem[Mill\'an et~al.(2009)Mill\'an, Rosolen, and Arroyo]{millan}
Mill\'an, D., Rosolen, A., and Arroyo, M.
\newblock Numerical integration by using local-node gauss-hermite cubature.
\newblock 01 2009.

\bibitem[Murphy(2012)]{mlBook}
Murphy, K.~P.
\newblock \emph{Machine Learning: A Probabilistic Perspective}.
\newblock The MIT Press, 2012.

\bibitem[Nagel \& Sudret(2016)Nagel and Sudret]{nagelSudret}
Nagel, J.~B. and Sudret, B.
\newblock Spectral likelihood expansions for bayesian inference.
\newblock \emph{Journal of Computational Physics}, 309\penalty0 (15):\penalty0
  267--294, 2016.
\newblock \doi{https://doi.org/10.1016/j.jcp.2015.12.047}.

\bibitem[Olver et~al.(2010)Olver, Lozier, Boisvert, and Clark]{olver}
Olver, F. W.~J., Lozier, D.~M., Boisvert, R.~F., and Clark, C.~W.
\newblock \emph{Quadrature: Gauss-Hermite formula}.
\newblock Cambridge University Press, 2010.

\bibitem[Opper \& Saad(2001)Opper and Saad]{opper}
Opper, M. and Saad, D.
\newblock \emph{Advanced Mean Field Methods: Theory and Practice}.
\newblock Neural Information Processing series. 2001.

\bibitem[Peterson \& Anderson(1987)Peterson and Anderson]{peterson}
Peterson, C. and Anderson, J.~R.
\newblock A mean field theory learning algorithm for neural networks.
\newblock \emph{Complex Systems}, 1\penalty0 (5):\penalty0 995--1019, 1987.

\bibitem[Riesz(1922)]{riesz}
Riesz, M.
\newblock Sur le problème des moments et le théorème de parseval
  correspondant.
\newblock \emph{Acta Sci. Math. Szeged}, 1:\penalty0 209--225, 1922.

\bibitem[Saul \& Jordan(1995)Saul and Jordan]{saulJordan}
Saul, L. and Jordan, M.
\newblock Exploiting tractable substructures in intractable networks.
\newblock \emph{Advances in Neural Information Processing Systems}, 8, 1995.

\bibitem[Schmuland(1992)]{schmuland}
Schmuland, M.
\newblock Dirichlet forms with polynomial domain.
\newblock \emph{Math. Japonica}, 37\penalty0 (6):\penalty0 1015--1024, 1992.

\bibitem[Simon(1971)]{simonh}
Simon, B.
\newblock Distributions and their hermite expansions.
\newblock \emph{Journal of Mathematical Physics}, 12\penalty0 (1), 1971.

\bibitem[Simon(2005)]{simon}
Simon, B.
\newblock \emph{Orthogonal polynomials on the unit circle. Part 1 \& Part 2},
  volume~54.
\newblock American Mathematical Society, 2005.

\bibitem[Soelch et~al.(2016)Soelch, Bayer, Ludersdorfer, and van~der
  Smagt]{soelch}
Soelch, M., Bayer, J., Ludersdorfer, M., and van~der Smagt, P.
\newblock Variational inference for on-line anomaly detection in
  high-dimensional time series.
\newblock \emph{Anamoly Detection Workshop Paper, International Conference on
  Machine Learning (ICML)}, 2016.

\bibitem[Theis \& Hoffman(2015)Theis and Hoffman]{theis}
Theis, L. and Hoffman, M.~D.
\newblock A trust-region method for stochastic variational inference with
  applications to streaming data.
\newblock \emph{Proceedings of the 32nd International Conference on Machine
  Learning}, 37:\penalty0 2503--2511, 2015.

\bibitem[vanZandt(2017)]{zandt}
vanZandt, J.
\newblock Efficient cubature rules.
\newblock \emph{Electronic Transactions on Numerical Analysis}, 51, 2017.

\bibitem[Vinck et~al.(2012)Vinck, Battaglia, Balakirsky, Vinck, and
  Pennartz]{vinck}
Vinck, M., Battaglia, F.~P., Balakirsky, V.~B., Vinck, A. J.~H., and Pennartz,
  C. M.~A.
\newblock Estimation of the entropy based on its polynomial representation.
\newblock \emph{Phys. Rev. E}, 85\penalty0 (5), 2012.
\newblock \doi{https://doi.org/10.1103/PhysRevE.85.051139}.

\bibitem[Walter(1977)]{walter}
Walter, G.~G.
\newblock Properties of hermite series estimation of probability density.
\newblock \emph{Annals of Statistics}, 5\penalty0 (6):\penalty0 1258--1264,
  1977.

\bibitem[Waterhouse et~al.(1996)Waterhouse, MacKay, and Robinson]{waterhouse}
Waterhouse, S., MacKay, D., and Robinson, T.
\newblock Bayesian methods for mixtures of experts.
\newblock \emph{Neural Information Processing Systems}, 1996.

\end{thebibliography}
\bibliographystyle{icml2023}

%%%%%%%%%%%%%%%%%%%%%%%%%%%%%%%%%%%%%%%%%%%%%%%%%%%%%%%%%%%%%%%%%%%%%%%%%%%%%%%
%%%%%%%%%%%%%%%%%%%%%%%%%%%%%%%%%%%%%%%%%%%%%%%%%%%%%%%%%%%%%%%%%%%%%%%%%%%%%%%
% APPENDIX
%%%%%%%%%%%%%%%%%%%%%%%%%%%%%%%%%%%%%%%%%%%%%%%%%%%%%%%%%%%%%%%%%%%%%%%%%%%%%%%
%%%%%%%%%%%%%%%%%%%%%%%%%%%%%%%%%%%%%%%%%%%%%%%%%%%%%%%%%%%%%%%%%%%%%%%%%%%%%%%

\appendix
\onecolumn

\section{Appendix}

\subsection{Notations and Dimensions for OPAA} \label{appendixDi}
For clarity, we provide a list of the notations used in this paper and their dimensions. We hope it will help the reader navigate through the details.
\begin{itemize}
	\item There are $N$ \textbf{latent variables}. The order of Gauss--Hermite quadrature is $\Gamma$.
	\item The $N$ latent variables are $\theta_1, \theta_2, \dots, \theta_N$. Sometimes, they are expressed as a vector $\theta$ or $\theta_{1:N}$ in $\Rn$.
	
	\item A \textbf{multi-index} $\tau = (i_1, i_2, \dots, i_N)$ is a vector of length $N$, where the indices $i_k$ are all non-negative integers.
	\item The \textbf{index set} $I$ (see (\ref{indexsetDef})) is the Cartesian product $[1, 2, \dots, \Gamma]^N$ which contains $\Gamma^N$ elements.
	\item Each element of the index set $I$ is called a \textbf{grid-index}, $(j) = (j_{1}, j_2, \dots,  j_N)$. It is a vector of length $N$, and each entry $j_k$ is an integer between $1$ and $\Gamma$ (inclusive).
	\item For each grid-index $(j)$ in $I$, there corresponds a \textbf{quadrature weight} $\tilde{w}^{(j)}$ and a \textbf{node} $\tilde{r}^{(j)}$ (defined in \\eqref{wtildeDef} and \eqref{rtildeDef} respectively).
	\item In higher dimensions, each quadrature weight $\tilde{w}^{(j)} \in \R$ is a real number from the multiplication of $N$ weights, $\tilde{w}_{j_1}, \dots, \tilde{w}_{j_N}$.
	\item In higher dimensions, each quadrature node $\tilde{r}^{(j)}$ is a vector of length $N$, namely, $(\tilde{r}_{j_1}, \dots, \tilde{r}_{j_N})$.
\end{itemize}

%You can have as much text here as you want. The main body must be at most $8$ pages long.
%For the final version, one more page can be added.
%If you want, you can use an appendix like this one, even using the one-column format.
\subsection{Hermite Functions} \label{secHF}
\begin{figure}[ht]
	\vskip 0.2in
	\begin{center}
		\centerline{\includegraphics[width=0.7\columnwidth]{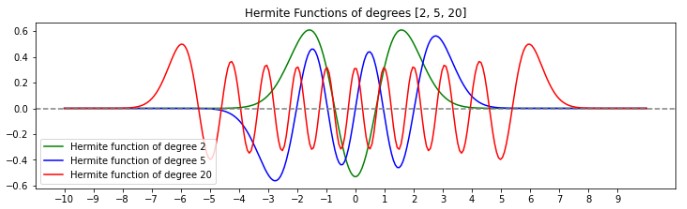}}
		\caption{Plots of Hermite functions for $n=2, 5, 10$.}
		\label{hermitePlot}
	\end{center}
	\vskip -0.2in
\end{figure}

To understand intuitively why the integral in equation \eqref{atauDef} should converge,  observe that it can be rewritten as
\be
a_\tau := \int_{\Rn} P(\theta)^{1/2} \left(\prod_{j=1}^N  h_{i_j}(\theta_j) e^{-\theta_j^2/2}\right) d\theta_{1:N}  = \int_{\Rn}  P(\theta)^{1/2}  \prod_{j=1}^N \psi_{i_j}(\theta_j) d\theta_{1:N}
\ee for $\tau = (i_1, i_2, \dots, i_N)$, where
\ben
\psi_n(x) := h_n(x) e^{-x^2/2}, n \geq 0
\label{hermiteFunctions}
\een are known as the \textbf{Hermite functions}. Properties of Hermite functions, such as their asymptotics, have been a subject of interest for statisticians and physicists alike \citep{walter, simonh}. The most relevant one for this paper could be the Cramer's Inequality \citep{hille} for Hermite functions, 
\be
\psi_n(x) \leq K \pi^{-1/4} \text{ on } \R,
\ee where $K < 1.09$ is a constant. Furthermore, the $n$-th Hermite function $\psi_n(x)$ satisfies the differential equation\footnote{This equation is equivalent to the Schrödinger equation for a harmonic oscillator in quantum mechanics.}
\be
\psi_n''(x) + (2n+1-x^2) \psi_n(x)= 0.
\ee More precisely, the asymptotics of the Hermite function can be described by the formula below,
\be
\psi_n(x) = 2^{1/4 n^{-1/12}} \left( Ai(\sqrt{2} n^{1/6} t) + O(n^{-2/3}) \right) \text{ for } |x| = \sqrt{2n+1} + t , t \geq 0,
\ee where $Ai$ is the Airy function of the first kind. In particular, Hermite functions decay very rapidly outside the interval $[-\sqrt{2n+1}, \sqrt{2n+1}]$. Shown in Figure \ref{hermitePlot} are the plots of Hermite functions of degrees 2, 5, and 10 respectively. 

\subsection{Example: Gaussian Mixture Model (GMM).} \label{mixedGaussian}
We consider the Gaussian Mixture Model with $K_0$ clusters centered at $\mu_{1:K_0}$ respectively, where
\be
\mu_k \sim \mathcal{N}(0, \sigma_\mu^2), 1 \leq k \leq K_0.
\ee
The probability density of the model is given by
\be
p(x; \mu_{1:K_0}) \sim \sum_{k=1}^{K_0}  \pi_k N(\mu_{z_k}, \sigma_x^2),
\ee where $\pi = (\pi_1, \pi_2, \dots, \pi_{K_0})$ is a random variable of dimension with only one non-zero entry of value $1$ in its output. Here is an example to illustrate this.

\begin{figure}[ht]
	\vskip 0.2in
	\begin{center}
		\centerline{\includegraphics[width=0.7\columnwidth]{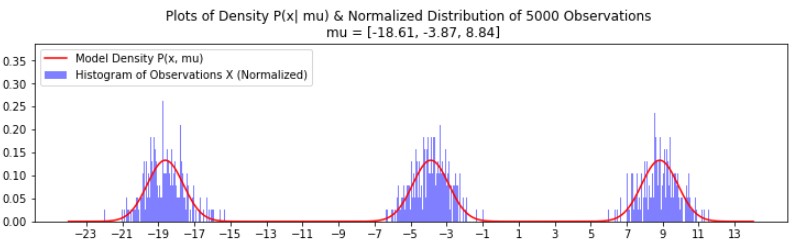}}
		\caption{Example of a Mixed Gaussian Model.}
		\label{gmmExamplebasic}
	\end{center}
	\vskip -0.2in
\end{figure}

We ran this experiment: first, we sampled $N_0 = 3$ points, $\mu \sim N(0, 10)$ and obtained $\mu_1 = -18.61$, $\mu_2 = 3.81$ and $\mu_3 = 8.84$. Then we generated $n=1000$ samples by first randomly selecting an integer $i$ from $[1, 2, 3]$, and then drawing $x\sim N(\mu_i, 1)$. Figure \ref{gmmExamplebasic} presents a plot of the joint distribution $p(x, \mu_1, \mu_2, \mu_3)$ of this particular experiment, alongside with a normalized histogram of these $1000$ samples.

In general, we are interested in the inverse problem of approximating the posterior
\ben
P(\mu| x_{1:n}): \Rn \mapsto \R
\label{posteriorGMM}
\een as a function of the latent variables $\mu \in \Rn$ given the observations $x_{1:n}$. Observe that the joint probability density function is given by
\ben
P(\mu_{1:K_0}, x_{1:n}) = \prod_{k=1}^{K_0} p(\mu_k)  \prod_{i=1}^n p(x_i |  \mu_{1:K_0}) .
% = \prod_{k=1}^{K_0} \frac{e^{\frac{-\mu_k^2} {2}}}{\sqrt{2\pi}} \prod_{i=1}^n \left( \ds \frac{1}{ K_0 \sqrt{2\pi}} \sum_{k=1}^{K_0} e^{\frac{-(x_i-\mu_k)^2}{2}} \right)
\label{jointFormula}
\een

To obtain the posterior in (\ref{posteriorGMM}), one needs the normalizing weight $P(x_{1:n})$, which requires us to sum (\ref{jointFormula}) in $k$ and integrate in $\mu_{1:K_0}$ . First, note that for any one sample $x$,
\ben
P(x| \mu_{1:K_0}) = \ds \sum_{k=1}^{K_0} p(x, z_k| \mu_{1:K_0}) % \prod_{k=1}^{K_0}
= \sum_{k=1}^{K_0} p(\mu_k | \mu_{1:K_0}) p(x, \mu_k) = \frac{1}{K_0}\sum_{k=1}^{K_0}  p(x, \mu_k) .
\label{cond1}
\een

Then we need to integrate (\ref{cond1}) against $d\mu_{1:K_0}$. That results in the following formula
\ben
P(x_{1:n}) = \int_{\mu_{1:{K_0}}} \prod_{k=1}^{K_0} p(\mu_k)  \prod_{i=1}^n \left(\sum_{k=1}^{K_0} \ds \frac{1}{ K_0} p(x_i, \mu_k) \right) d\mu_{1:K_0} .
%\prod_{k=1}^{K_0} \frac{e^{\frac{-\mu_k^2} {2}}}{\sqrt{2\pi}} \prod_{i=1}^n \left( \ds \frac{1}{ K_0 \sqrt{2\pi}} \sum_{k=1}^{K_0} e^{\frac{-(x_i-\mu_k)^2}{2}} \right)
\label{evidenceFormula}
\een

While it may be possible to compute (\ref{evidenceFormula}) directly, the computation is far from straightforward. Furthermore, there are $K_0^n$ terms, making the computations extremely expensive as $n$ increases.

%\subsubsection{Computation Times} \label{compTimes}
%Here are the computation times of all the required quadrature weights on a 6-core CPU. Recall that $N$ is the number of latent variables, $\Gamma$ is the degree of approximation, $I$ is the index set (defined in (\ref{indexsetDef})). Our numerical experiments have shown that $\Gamma \leq 8$ seems to be sufficient.

\end{document}